\documentclass{article} 
\usepackage{iclr2019_conference_arxiv,times}

\usepackage{amsmath,amsfonts,bm}

\def\eqref#1{equation~\ref{#1}}

\def\1{\bm{1}}

\def\rx{{\textnormal{x}}}
\def\ry{{\textnormal{y}}}

\def\rvx{{\mathbf{x}}}

\def\vh{{\bm{h}}}

\def\mC{{\bm{C}}}

\def\mE{{\bm{E}}}

\DeclareMathAlphabet{\mathsfit}{\encodingdefault}{\sfdefault}{m}{sl}
\SetMathAlphabet{\mathsfit}{bold}{\encodingdefault}{\sfdefault}{bx}{n}

\usepackage{hyperref}
\usepackage{url}
\hypersetup{breaklinks=true}

\usepackage{amsmath,amssymb}
\usepackage[utf8]{inputenc}
\usepackage[arabic,french,spanish,russian,english]{babel}
\usepackage{CJKutf8}
\usepackage{times}
\usepackage{latexsym}
\usepackage{booktabs}
\usepackage{subcaption}
\usepackage{graphicx}
\usepackage{tabularx}
\usepackage{diagbox}

\usepackage{soul}
\usepackage{color}

\definecolor{lightblue}{rgb}{0.5,0.5,1}
\definecolor{lightred}{rgb}{1,0.5,0.5}

\usepackage{todonotes} 

\usepackage{multirow}
\usepackage[normalem]{ulem}
\usepackage{enumitem}

\usepackage{url}

\usepackage{array}
\newsavebox{\RTLbox}
\newcolumntype{R}{>{\begin{lrbox}{\RTLbox}}r<{\end{lrbox}\AR{\unhbox\RTLbox}}}
\newcolumntype{L}{>{\begin{lrbox}{\RTLbox}}l<{\end{lrbox}\AR{\unhbox\RTLbox}}}

\newcolumntype{H}{>{\setbox0=\hbox\bgroup}c<{\egroup}@{}}

\usepackage{lingmacros}

\newcommand{\hh}{\mathbf{h}}

\title{Identifying and Controlling Important \\ Neurons in Neural Machine Translation }

\author{  Anthony Bau$^{1}$\thanks{Equal contribution} \hfill Yonatan Belinkov$^{1*}$ \hfill Hassan Sajjad$^{2}$  \\  \textbf{Nadir Durrani}$^{2}$  \hfill  \textbf{Fahim Dalvi}$^{2}$ \hfill \textbf{James Glass}$^{1}$ \\\\
  $^{1}$MIT Computer Science and Artificial Intelligence Laboratory \\
  $^{2}$Qatar Computing Research Institute -- HBKU \\ 
  \texttt{\{abau,belinkov,glass\}@mit.edu} \\
  \texttt{\{hsajjad,ndurrani,faimaduddin\}@qf.org.qa} \\
}

\iclrfinalcopy 
\begin{document}

\maketitle

\begin{abstract}
Neural machine translation (NMT) models learn representations containing substantial linguistic information. However, it is not clear if such information is fully distributed or if some of it 
can be attributed to individual neurons. 
We develop 
unsupervised methods for discovering important neurons in NMT models. Our methods rely on the intuition that different models  
learn similar properties, and do not require any costly external supervision. 
We show experimentally that translation quality depends on the discovered neurons, 
and find that many of them capture common linguistic phenomena. 
Finally, we show how to control NMT translations in predictable ways, by modifying  activations of  individual neurons.   

\end{abstract}

\section{Introduction}

 Neural machine translation (NMT) systems achieve state-of-the-art results by learning from large amounts of example translations,  typically 
 without additional linguistic information. 
 Recent studies have shown that representations learned by NMT models contain a non-trivial amount of linguistic information on multiple levels: morphological~\citep{P17-1080}, syntactic~\citep{shi-padhi-knight:2016:EMNLP2016}, and semantic~\citep{
 Hill2017}. 
 These studies use trained NMT models to generate feature representations for words,  
 and use these representations to predict certain linguistic properties. This approach has two main limitations. First, it targets the whole vector representation and fails to analyze individual dimensions in the vector space. In contrast, previous work found
 meaningful individual neurons 
 in computer vision~\citep[among others]{zeiler2014visualizing,zhou2015cnnlocalization,netdissect2017} and in a few NLP tasks~\citep{karpathy2015visualizing,radford2017learning,qian-qiu-huang:2016:EMNLP2016}. 
 Second, these methods require external supervision in the form of linguistic annotations. They are therefore limited by available annotated data and 
 tools.

 In this work, we make initial progress towards addressing these limitations by developing unsupervised methods for analyzing the contribution of \textit{individual neurons} to NMT models. We aim to answer the following questions: 
 \begin{itemize}[leftmargin=30pt,itemsep=0pt,topsep=3.5pt]
 \item How important are individual neurons 
 for obtaining high-quality translations? 
 \item Do individual neurons in NMT models contain interpretable linguistic information? 
 \item Can we control MT output by intervening in the representation at the individual neuron level? 
 \end{itemize}

To answer these questions, we develop several unsupervised methods for ranking neurons according to their importance to an NMT model. Inspired by work in machine vision~\citep{li2016convergent}, we hypothesize that different NMT models 
learn similar properties, and therefore similar important neurons should emerge in different models. 
To test this hypothesis, we map neurons between pairs of trained NMT models using several methods: correlation analysis, regression analysis, and SVCCA, a recent method combining singular vectors and canonical correlation analysis~\citep{NIPS2017_7188}.
Our mappings yield lists of candidate neurons containing shared information across models. We then evaluate whether these 
neurons carry important information to the NMT model 
by masking their activations during testing. We find that highly-shared neurons impact translation quality much more than unshared neurons, affirming our hypothesis that {\it shared information matters}.

Given the list of important neurons, we then investigate what linguistic properties they 
capture, both qualitatively by visualizing neuron activations and quantitatively by performing supervised classification experiments. 
We were able to identify neurons corresponding to several linguistic phenomena, including morphological and syntactic properties.

Finally, we 
test whether intervening in the representation at the individual neuron level can help \textit{control the translation}.  
We demonstrate the ability to control NMT  
translations on three linguistic properties---tense, number, and gender---to varying degrees of success. 
This 
sets the ground for controlling NMT in desirable ways, 
potentially reducing system bias to properties like 
gender.

Our work indicates that not all information is distributed in NMT models, and that many human-interpretable grammatical and structural properties are captured by individual neurons. 
Moreover, modifying the activations of individual neurons allows controlling the translation output according to specified linguistic properties.
The methods we develop here are task-independent and can be used for 
analyzing neural networks in other 
tasks. 
More broadly, our work contributes to the localist/distributed debate in neural cognitive science~\citep{doi:10.1080/09540091.2011.587505} by investigating the important case of neural machine translation.

\section{Related Work} \label{sec:related-work}
Much recent work has been concerned with analyzing neural representations of linguistic units, such as word embeddings~\citep{kohn:2015:EMNLP,qian-qiu-huang:2016:P16-11}, sentence embeddings~\citep{adi2016fine,Ganesh:2017:IST:3110025.3110083,brunner2018natural}, and 
NMT representations at different linguistic levels: morphological~\citep{P17-1080}, syntactic~\citep{shi-padhi-knight:2016:EMNLP2016}, and semantic~\citep{Hill2017}. 
These studies follow a common methodology of evaluating learned representations on external supervision by training classifiers or measuring other kinds of correlations. 
Thus they are limited to the available supervised annotation. 
In addition, these studies also do not typically consider individual dimensions. 
In contrast, we propose intrinsic unsupervised methods for detecting important neurons based on correlations between independently trained models. A similar approach was used to analyze vision networks~\citep{li2016convergent}, but to the best of our knowledge these ideas were not used to study NMT or other NLP models before. 

In computer vision, individual neurons were shown to capture meaningful information~\cite{zeiler2014visualizing,zhou2015cnnlocalization,netdissect2017}. 
Even though some doubts were cast on the importance of individual units~\citep{s.2018on}, recent work stressed their contribution to predicting specific object classes via masking experiments similar to ours~\citep{zhou2018revisiting}. 
A few studies analyzed individual neurons in NLP. For instance, neural language models learn specific neurons that activate on brackets~\citep{karpathy2015visualizing},  sentiment~\citep{radford2017learning}, and length~\citep{qian-qiu-huang:2016:EMNLP2016}. Length-specific neurons were also found in NMT 
\citep{D16-1248}, but generally not much work has been devoted to analyzing individual neurons in NMT. 
We aim to address this gap.

\section{Methodology} \label{sec:methods}

Much recent 
work on analyzing NMT relies on 
supervised learning,  
where NMT representations are used as features for predicting linguistic annotations (see Section~\ref{sec:related-work}).
However, such annotations may not be available, or 
constrain the analysis to a particular scheme. 

Instead, we propose to use different kinds of correlations between neurons from different models as a measure of their importance. Suppose we have $M$ such models
and 
let $\hh_t^m[i]$ denote the activation of the $i$-th neuron in the encoder of the $m$-th model for the $t$-th word.\footnote{
We only consider neurons from the top 
layer, although the approach can also be applied to other layers. 
} 
These may be models from different training epochs,  trained with different random initializations or 
datasets, 
or even 
different architectures---all 
realistic scenarios that researchers often experiment with. 
Let $\rx_i^m$ denote a random variable corresponding to the $i$-th neuron in the $m$-th model.
$\rx_i^m$ maps words to their neuron activations: $ \rx_i^m : t \mapsto \hh_t^m[i] $. 
Similarly, let $\rvx^m$ denote a random vector corresponding to the activations of all neurons in the $m$-th model: $ \rvx^m : t \mapsto \vh_t^m $.

\begin{figure}[t]
\includegraphics[width=\linewidth]{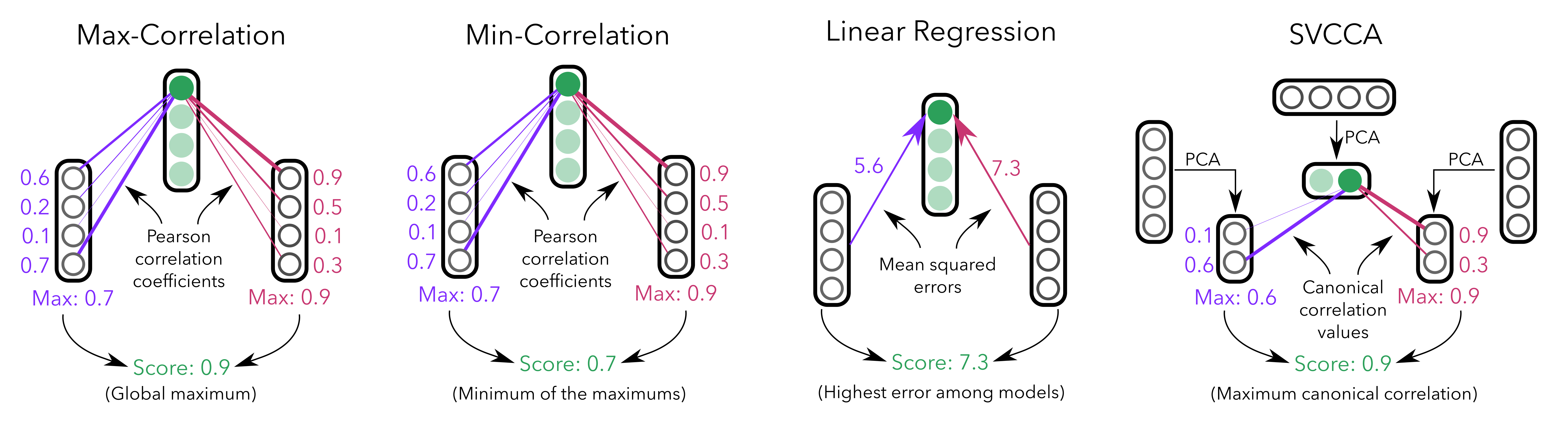}
\caption{An illustration of the correlation methods methods, showing how to compute the score for one neuron using each of the methods. Here the number of models is $M=3$.}
\label{fig:methods}
\end{figure}

We consider four 
methods for \textit{ranking} neurons, based on 
correlations between pairs of models.
Our hypothesis is that different NMT models 
learn similar properties, and therefore similar important neurons 
emerge in different models, akin to neural vision models~\citep{li2016convergent}. 
 Our methods capture different levels of localization/distributivity, as 
 described next. 
 See Figure~\ref{fig:methods} for 
 illustration.

\subsection{Unsupervised correlation Methods} \label{sec:methods-unsupervised}

\paragraph{Maximum correlation}
The maximum correlation (\texttt{MaxCorr}) 
of neuron $\rx_i^m$ looks for the highest correlation with any neuron in all other models:
\begin{equation} 
\texttt{MaxCorr}(\rx_i^m) = \max_{j, m' \neq m} |\rho(\rx_i^m, \rx_j^{m'})|
\end{equation} 
where $\rho(\rx,\ry)$ is the Pearson correlation coefficient between $\rx$ and $\ry$.  
We then rank the neurons in model $m$ according to their \texttt{MaxCorr} score. We repeat this procedure for every model $m$. 
This score looks for neurons that capture properties that emerge strongly in two separate models.  

\paragraph{Minimum correlation}
The minimum correlation (\texttt{MinCorr}) of neuron $\rx_i^m$ looks for the neurons most correlated with $X_i^m$ in each of the other models, 
but  
selects the one with the lowest correlation:
\begin{equation} 
\texttt{MinCorr}(\rx_i^m) = \min_{m' \neq m} \max_j |\rho(\rx_i^m, \rx_j^{m'})|
\end{equation} 
Neurons in model $m$ are ranked according to their \texttt{MinCorr} score. 
This 
tries  
to find neurons that are well correlated with many other models, even if they are not 
the overall 
most correlated ones.  

\paragraph{Regression ranking}
We perform linear regression (\texttt{LinReg})
from the full representation of another model $\rvx^{m'}$ to the neuron $\rx_i^m$. 
Then we rank neurons by the regression mean squared error. 
This attempts to find neurons whose information might be distributed in other models. 

\paragraph{SVCCA}
Singular vector canonical correlation analysis (\texttt{SVCCA}) is a recent 
method for analyzing neural networks~\citep{NIPS2017_7188}. In our implementation, we
perform 
PCA on each model's representations $\rvx^m$ and take enough dimensions to account for 99\% of the variance. For each pair of models, 
we obtain the canonically correlated basis, and 
rank the basis directions by their CCA 
coefficients. 
This attempts to capture information that may 
be distributed in 
less dimensions than the whole representation. In this case we get 
a ranking of directions, rather than  individual neurons.

\subsection{Verifying Detected Neurons}

We want to verify that 
neurons ranked highly by the unsupervised methods are indeed important for the NMT models. 
We consider quantitative and qualitative techniques for verifying their importance. 

\paragraph{Erasing Neurons}
We  
test importance of neurons by erasing some of them during translation. Erasure is 
a useful technique for analyzing neural networks~\citep{li2016understanding}. 
Given a ranked list of neurons $\pi$, where $\pi(i)$ is the rank of neuron $\rx_i$, 
we zero-out increasingly more 
neurons according to the ranking $\pi$, starting from either the top or the bottom of the list. Our hypothesis is 
that erasing neurons from the top 
would hurt translation performance more than erasing from the bottom.

Concretely, we first run the entire encoder as usual, then zero out specific neurons from all 
source hidden states $\{\hh_1, \dots, \hh_n\}$ before running the decoder. 
For \texttt{MaxCorr}, \texttt{MinCorr}, and \texttt{LinReg}, 
we zero out individual neurons. 
To erase $k$ directions found by \texttt{SVCCA}, we instead project the embedding $\mE$ (corresponding to all activations 
of a given model over a dataset) onto the space spanned by the non-erased directions:
$\mE' = \mE(\mC(\mC^{T}\mC)^{-1}\mC^T)$,
where $\mC$ is the CCA projection matrix with the  first or last $k$ columns removed. This corresponds to erasing from the top or 
bottom.

\paragraph{Supervised Verification}
While our focus is on unsupervised methods for finding important neurons, we also utilize supervision to verify our results. 
Since training a supervised classifier on every neuron is costly, we instead report simple metrics that can be easily computed. Specifically, we sometimes report the expected conditional variance of neuron activations conditioned on some property. In other cases we found it useful to estimate a Gaussian mixture model (GMM) for predicting a label and measure its prediction quality. We  obtain linguistic annotations with Spacy: 
\url{spacy.io}.

\paragraph{Visualization}
Interpretability of machine learning models remains elusive 
\citep{lipton2016mythos}, but visualizing can be an instructive technique. Similar to previous work analyzing neural networks in NLP~\citep{elman1991distributed,karpathy2015visualizing,kadar2016representation}, we visualize activations of neurons and observe interpretable behavior. We will illustrate this with example heatmaps below.

\section{Experimental Setup}
\paragraph{Data}

We use 
the United Nations (UN) parallel 
corpus~\citep{ZIEMSKI16.1195} for all experiments. We train models from  English to 5 
languages: Arabic, Chinese, French, Russian, and Spanish, as well as 
an English-English auto-encoder. For each target language, we train 3 models on different parts of the training set, each with 500K sentences. In total, we have 18 models. 
This setting allows us to 
compare models trained on the same language pairs but different training data, as well as 
models trained on different language pairs.
We evaluate on the official test set. 

\paragraph{MT training}
We train 500 dimensional 2-layer 
LSTM 
encoder-decoder models with attention~\cite{bahdanau2014neural}. 
In order to study both word and sub-word  
properties, we use 
a word representation based on a character convolutional neural network (charCNN) as input to both encoder and decoder, which was shown 
to learn morphology 
in language modeling and NMT~\citep{kim2015character,P17-1080}.\footnote{We used this representation 
rather than BPE sub-word units~\citep{P16-1162} to facilitate interpretability with respect to specific words. In the experiments, we report word-based results unless noted otherwise.} 
While we focus here on recurrent NMT, 
our approach can be applied to other models like the Transformer~\citep{NIPS2017_7181}, which we leave for future work.

\section{Results}
\subsection{Erasure Experiments}

\begin{figure}[t]
\centering
 \begin{subfigure}{0.24\linewidth}
 \includegraphics[width=\linewidth]{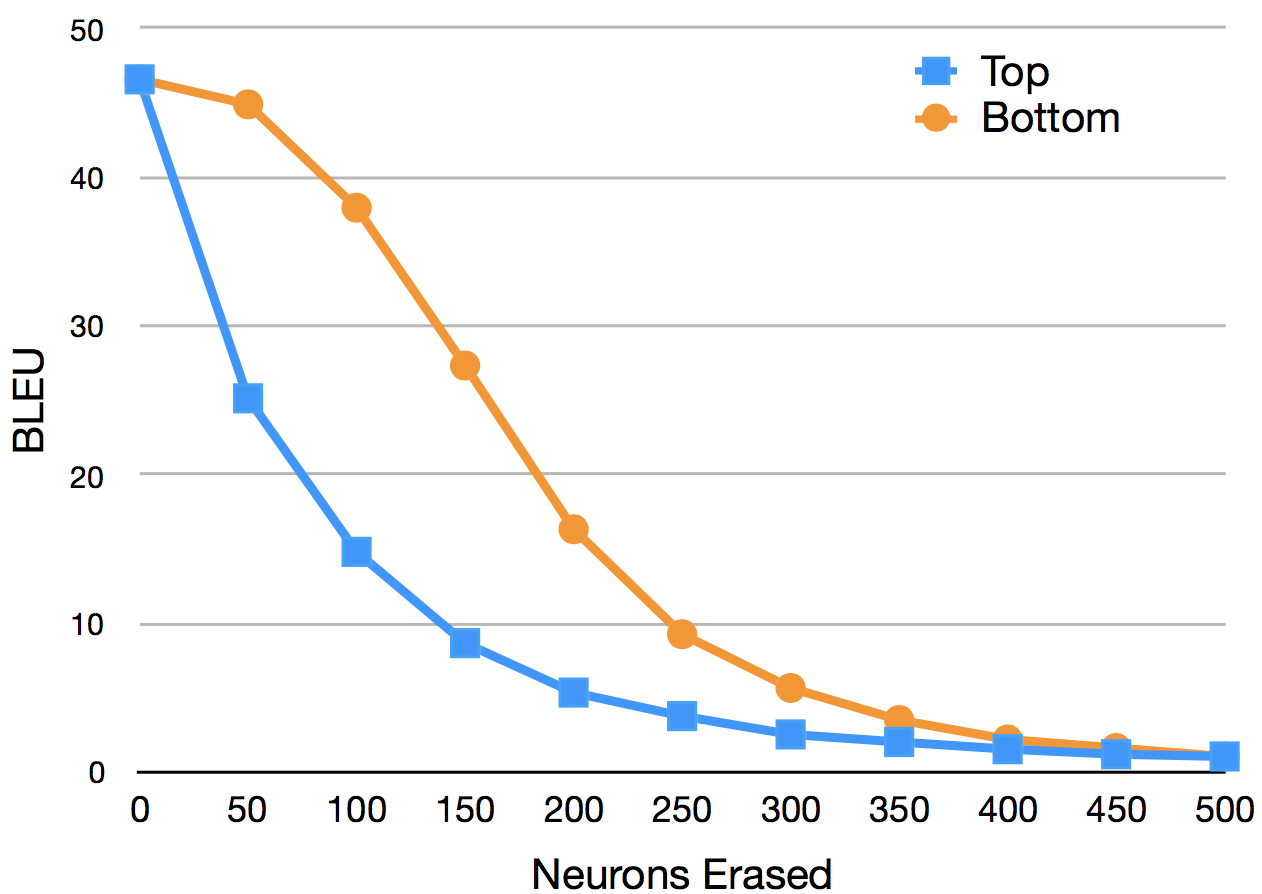}
 \caption{\texttt{MaxCorr}}
 \label{fig:erase-en-es-maxcorr}
 \end{subfigure}
 \begin{subfigure}{0.24\linewidth}
 \includegraphics[width=\linewidth]{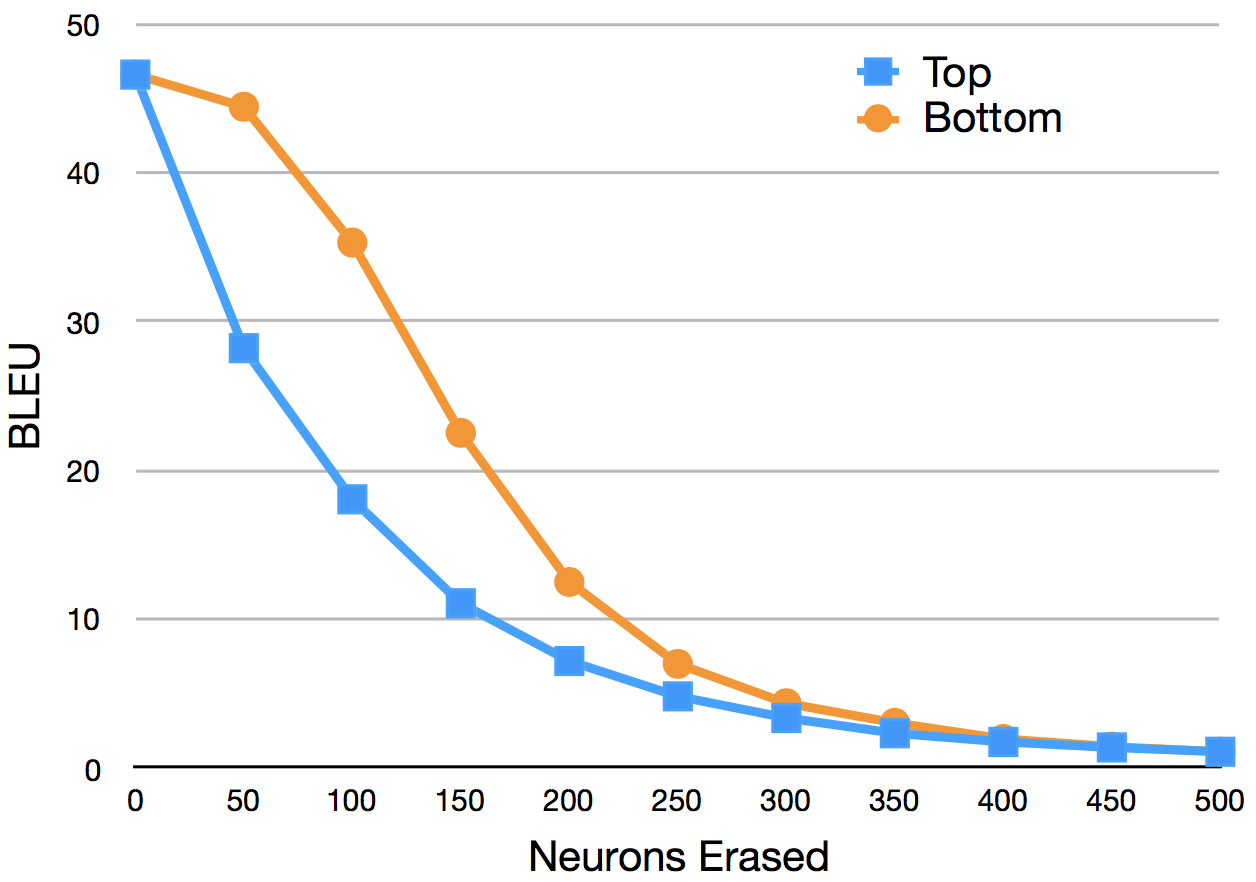}
 \caption{\texttt{MinCorr}}
 \label{fig:erase-en-es-maxcorr}
 \end{subfigure}
 \begin{subfigure}{0.24\linewidth}
 \includegraphics[width=\linewidth]{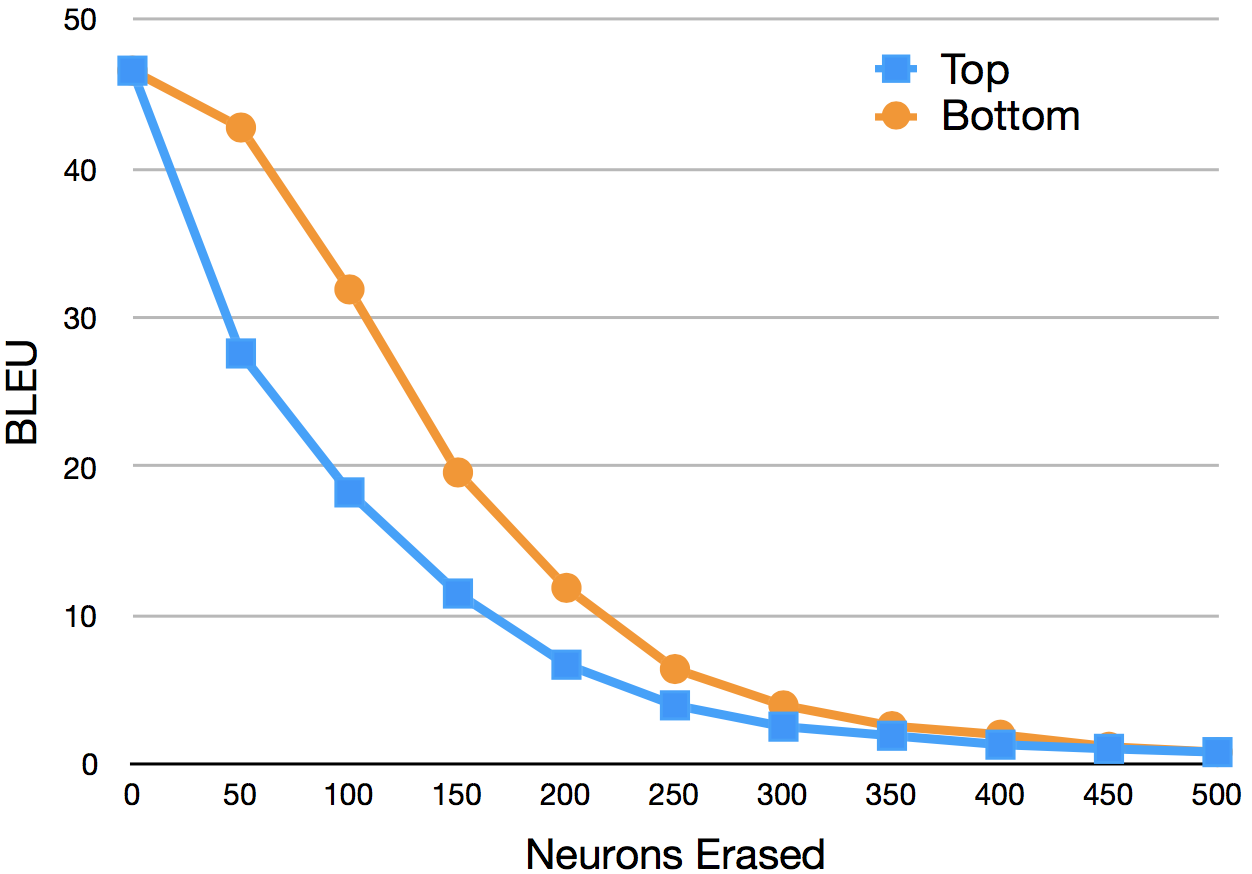}
 \caption{\texttt{LinReg}}
 \label{fig:erase-en-es-maxcorr}
 \end{subfigure}
 \begin{subfigure}{0.24\linewidth}
 \includegraphics[width=\linewidth]{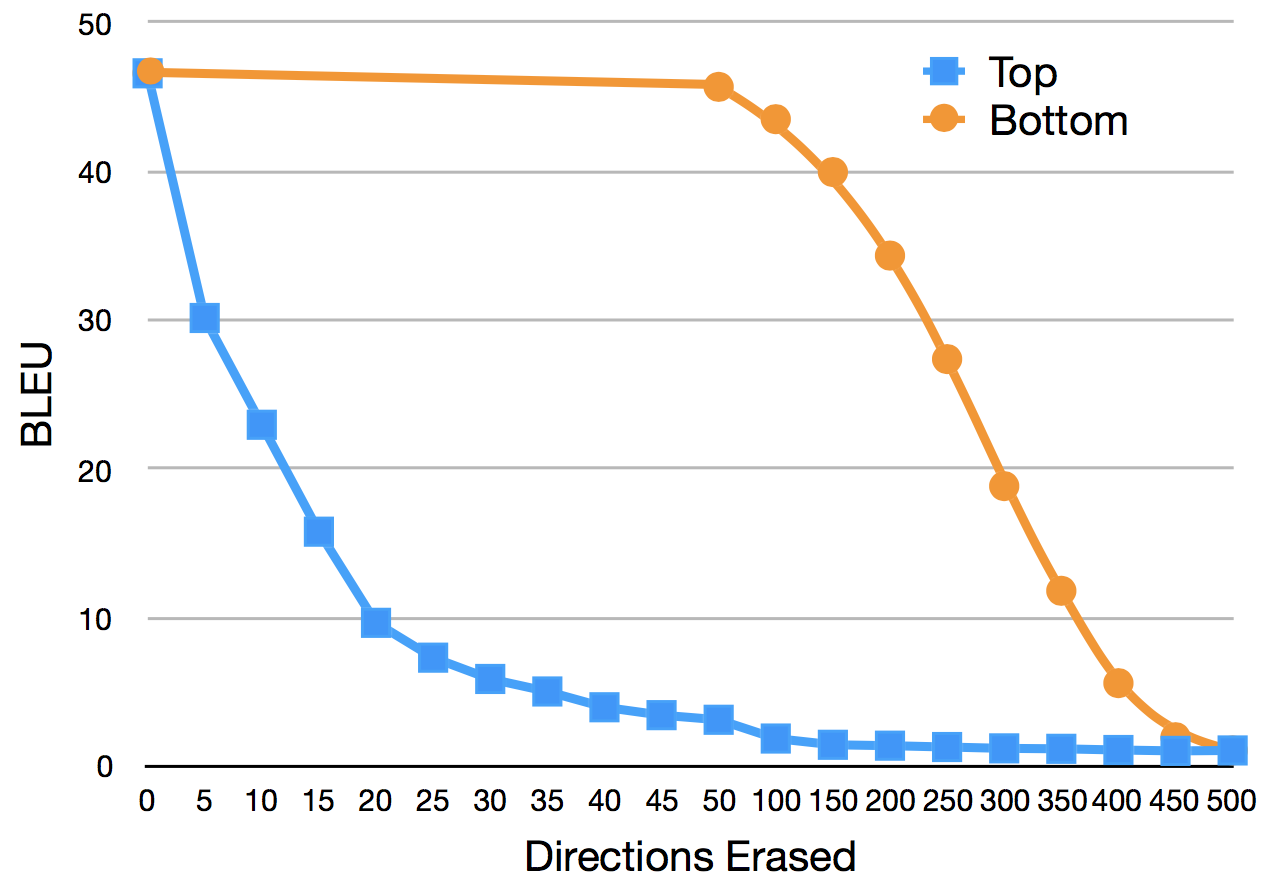}
 \caption{\texttt{SVCCA}}
 \label{fig:erase-en-es-svcca}
 \end{subfigure}
\caption{Erasing neurons (or \texttt{SVCCA} directions) from the top and bottom of the list of most important neurons (directions) ranked by different unsupervised methods, in an English-Spanish model.}
\label{fig:erase-all-methods}
\end{figure}

Figure~\ref{fig:erase-all-methods} shows erasure results 
using the  
methods from Section~\ref{sec:methods-unsupervised}, on an English-Spanish model. 
 For all four methods, erasing from the top hurts performance much more than erasing from the bottom. This confirms our hypothesis that 
 neurons ranked higher by our methods have a larger impact on translation quality. 
 Comparing erasure with different rankings, we find similar patterns with \texttt{MaxCorr}, \texttt{MinCorr}, and \texttt{LinReg}: erasing the top ranked 10\% (50 neurons) degrades BLEU by 15-20 points, while erasing the bottom 10\% neurons only hurts by \mbox{2-3} points. 
 In contrast, erasing \texttt{SVCCA} directions results in rapid degradation -- 15 BLEU point drop 
 when erasing 1\% (5) of the top directions, and  poor performance when erasing 10\% (50). This indicates that top 
 \texttt{SVCCA} directions capture very important  information in the model. 
 We analyze these 
 top neurons and directions in the next section, finding that top \texttt{SVCCA} directions focus  mostly on identifying 
 specific words.

 \begin{figure}[t]
\centering
 \begin{subfigure}{0.3\textwidth}
 \includegraphics[width=\linewidth]{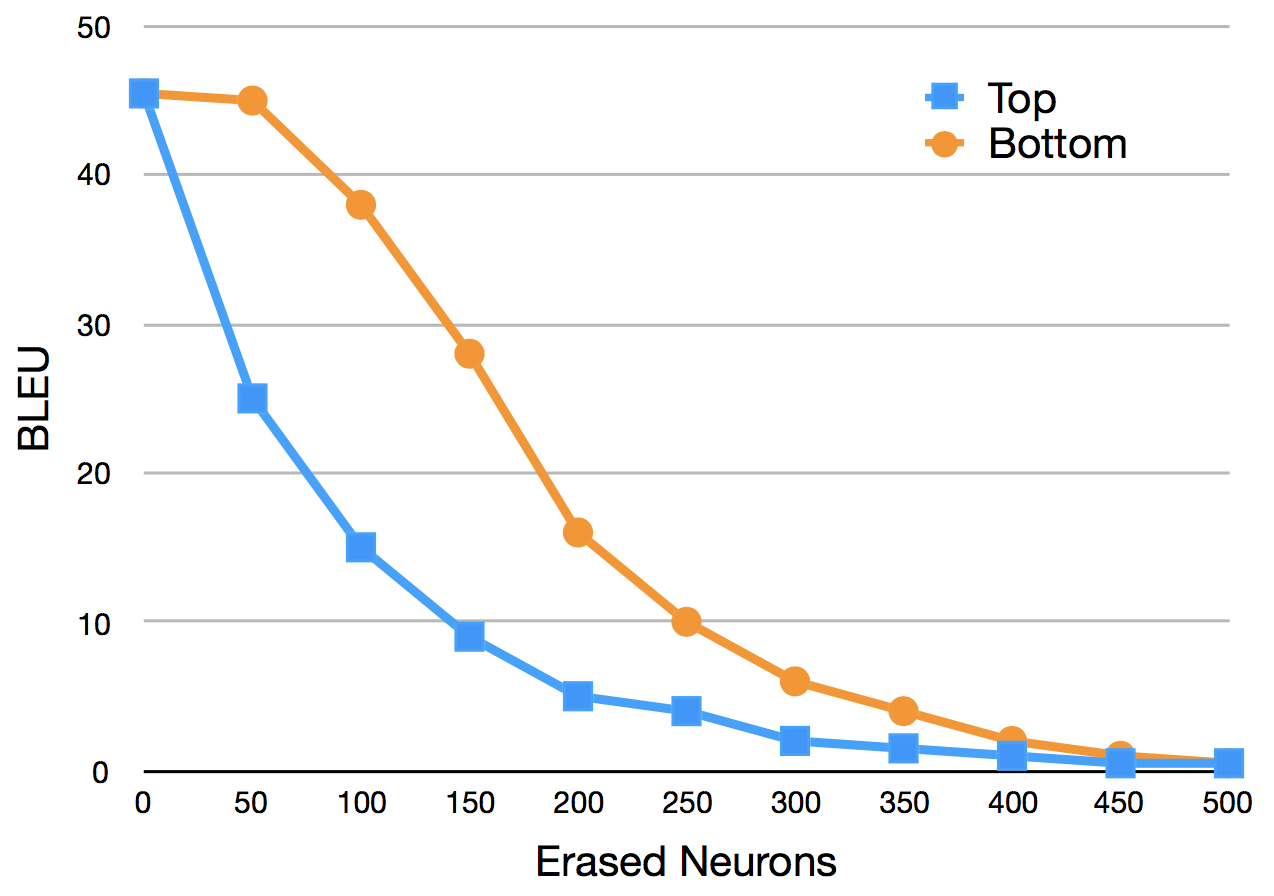}
 \caption{English-Spanish}
 \label{fig:erase-maxcorr-en-es-corpus}
 \end{subfigure}\hfill
 \begin{subfigure}{0.3\textwidth}
 \includegraphics[width=\linewidth]{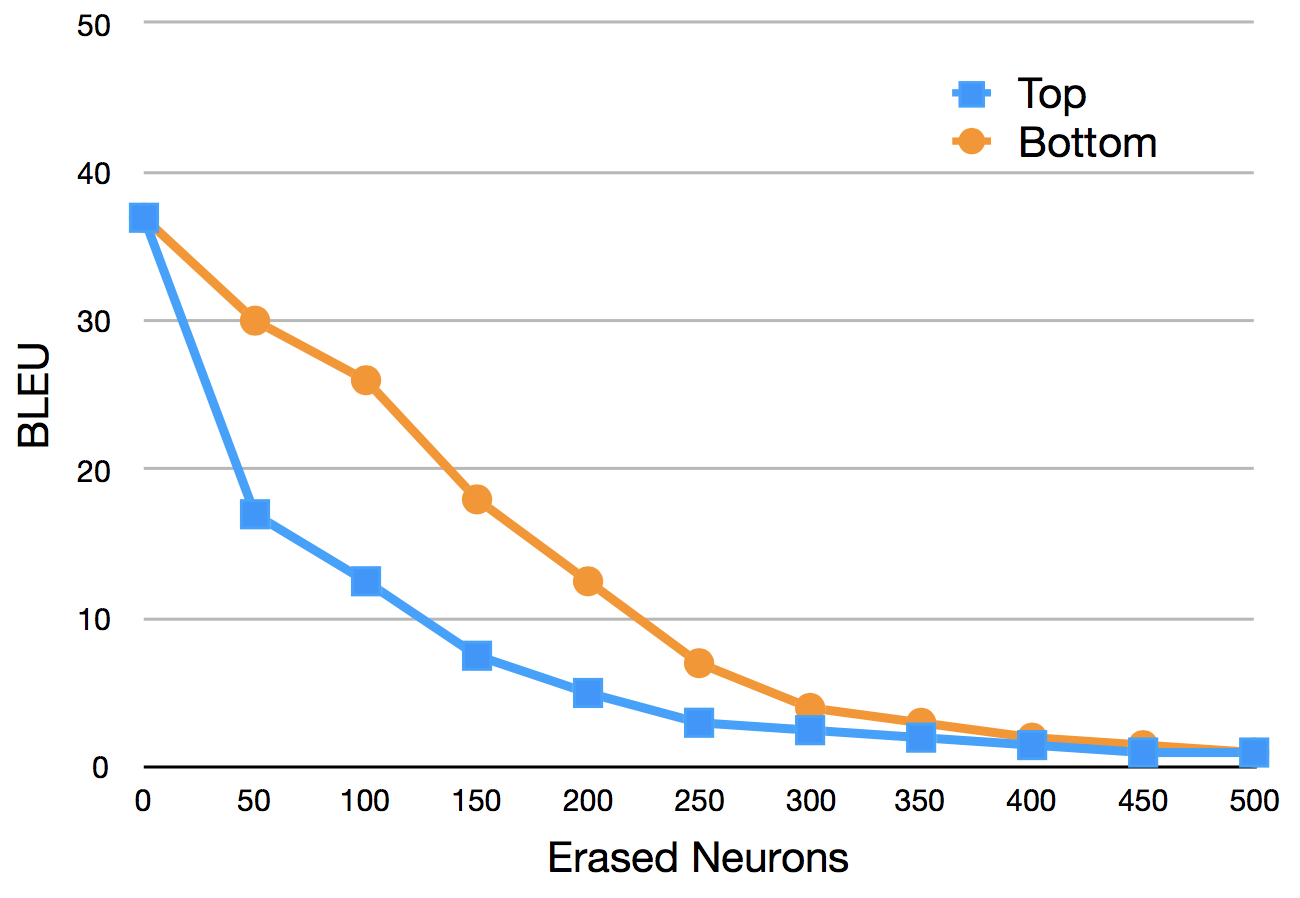}
 \caption{English-French}
 \label{fig:erase-maxcorr-en-fr-corpus}
 \end{subfigure}\hfill
 \begin{subfigure}{0.3\textwidth}
 \includegraphics[width=\linewidth]{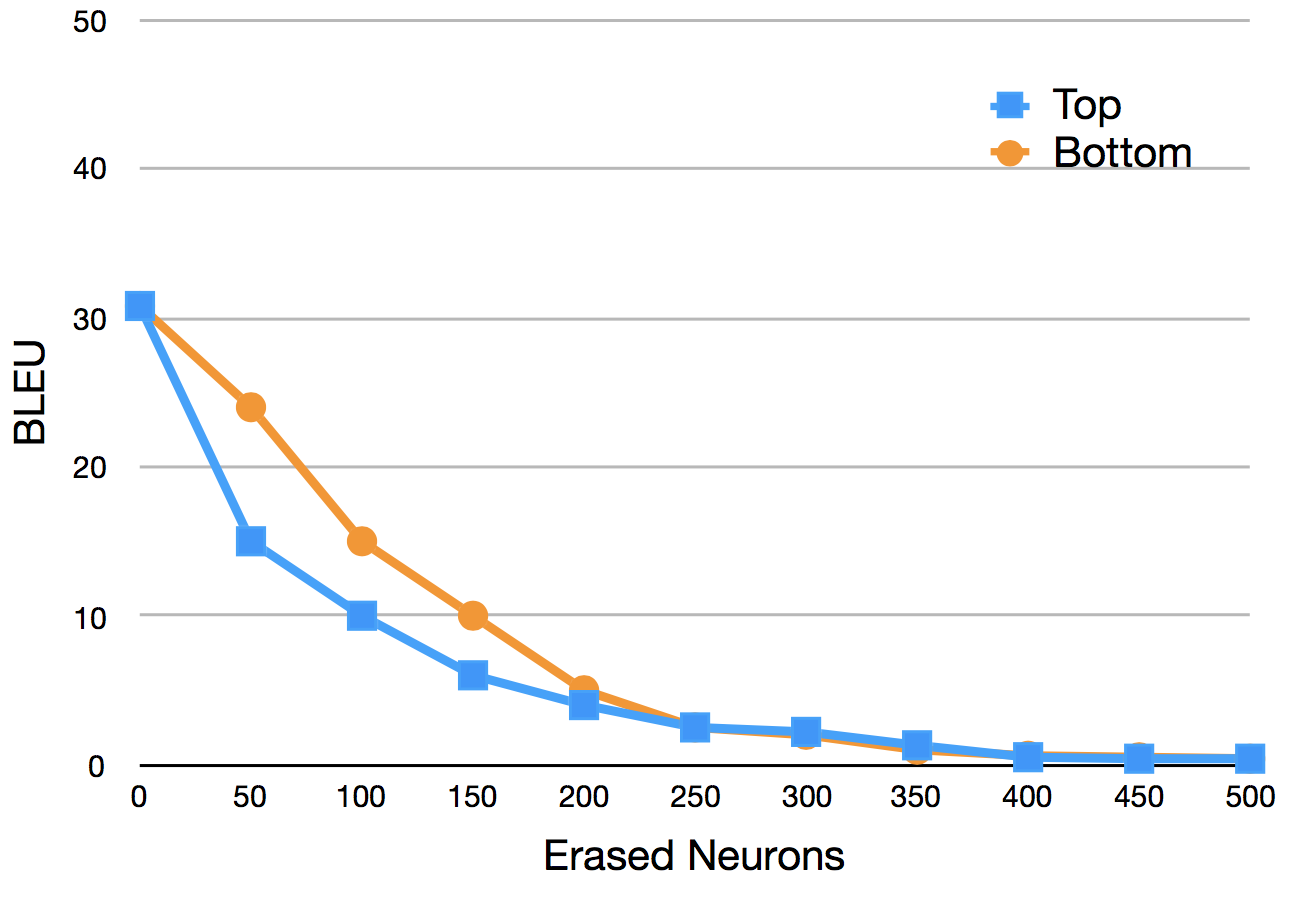}
 \caption{English-Chinese}
 \label{fig:erase-maxcorr-en-zh-corpus}
 \end{subfigure} 
\caption{
Erasing neurons from the top or bottom of the \texttt{MaxCorr} ranking in three language pairs.
} 
\label{fig:erase-maxcorr}
\end{figure}
 
 Figure~\ref{fig:erase-maxcorr}
shows the results of 
\texttt{MaxCorr}  
when erasing neurons from top and bottom,
using models trained on three language pairs. 
In all cases, 
erasing from the top hurts performance more
than erasing from the bottom.
We found similar trends with other language pairs and 
ranking methods.

\begin{table}[t]
\footnotesize
\centering
  \caption{Top 10 
  neurons (or \texttt{SVCCA} directions) in an English-Spanish model according to the four 
  methods, and the percentage of explained variance 
  by conditioning on position 
  or token identity. 
  } 
  \begin{tabularx}{\textwidth}{ X | X | X || X | X | X || X | X | X || X | X }
  \toprule
  \multicolumn{3}{c}{\texttt{MaxCorr}} &
  \multicolumn{3}{c}{\texttt{MinCorr}} &
  \multicolumn{3}{c}{\texttt{LinReg}} &
  \multicolumn{2}{c}{\texttt{SVCCA}} \\
  \midrule
  ID & Pos & Tok & ID & Pos & Tok & ID & Pos & Tok & Pos & Tok \\
  \midrule
 464 & \textbf{92\%} & 10\% 			& 342 & \textbf{88\%} & 7.9\%		& 464 & \textbf{92\%} & 10\% 		& \textbf{86\%} & 26\% \\
 342 & \textbf{88\%} & 7.9\%	 		& 464 & \textbf{92\%} & 10\% 		& 260 & 0.71\% & \textbf{94\%} 	& 1.6\% & \textbf{90\%}\\
 260 & 0.71\% & \textbf{94\%} 			& 260 & 0.71\% & \textbf{94\%} 		& 139 & 0.86\% & \textbf{93\%} 	& 7.5\% & \textbf{85\%}\\
 49 & 11\% & 6.1\% 					& 383 & \textbf{67\%} & 6.5\% 		& 494 & 3.5\% & \textbf{96\%} 	& 20\% & \textbf{79\%}\\
 124 & \textbf{77\%} & 48\% 			& 250 & \textbf{63\%} & 6.8\% 		& 342 & \textbf{88\%} & 7.9\% 	& 1.1\% & \textbf{89\%}\\
 394 & 0.38\% & 22\% 				& 124 & \textbf{77\%} & 47\% 		& 228 & 0.38\% & \textbf{96\%} 	& 10\% & \textbf{76\%}\\
 228 & 0.38\% & \textbf{96\%} 			& 485 & \textbf{64\%} & 10\% 		& 317 & 1.5\% & \textbf{83\%} 	& 30\% & \textbf{57\%}\\
 133 & 0.14\% & \textbf{87\%} 			& 480 & \textbf{70\%} & 12\% 		& 367 & 0.44\% & \textbf{89\%} 	& 24\% & \textbf{55\%}\\
 221 & 1\% & 30\% 					& 154 & \textbf{63\%} & 15\% 		& 106 & 0.25\% & \textbf{92\%} 	& 23\% & \textbf{60\%}\\ 
 90 & 0.49\% & 28\% 				& 139 & 0.86\% & \textbf{93\%} 		& 383 & \textbf{67\%} & 6.5\% 	& 18\% & \textbf{63\%}\\
 \bottomrule
  \end{tabularx}
  \label{top20}
\end{table}

\subsection{Evaluating Top Neurons}

What kind of information is captured by the neurons ranked highly by each of our ranking methods?
Previous work found specific neurons in NMT that capture position of words in the sentence~\citep{D16-1248}. Do our methods capture similar properties? Indeed, we found that many of the top neurons capture position. For instance, 
Table~\ref{top20} shows the top 10 
ranked neurons from an English-Spanish model 
according to each of the methods. 
The table shows the percent of variance in neuron activation that is eliminated by conditioning on position in the sentence, calculated over the test set. Similarly, it shows
the percent of explained variance by conditioning on the current token identity. 

We observe an interesting difference between the ranking methods. 
\texttt{LinReg} and especially \texttt{SVCCA}, which are both computed by using multiple neurons, tend to find information determined by the identity of the current token. \texttt{MaxCorr} and (especially) \texttt{MinCorr} tend to find position information. This suggests that information about the current token is often distributed in multiple neurons, which can be explained by the fact that 
tokens carry multiple kinds of linguistic information. 
In contrast, position is a fairly simple property that the NMT encoder can represent in a small number of neurons.

\subsection{Linguistically Interpretable Neurons}

Neurons that activate on specific tokens or capture position in the sentence are important, as shown in the previous section. But they are less interesting from the perspective of capturing language information. In this section, we investigate several linguistic properties by measuring predictive capacity and visualizing neuron activations.  The supplementary material discusses more properties.

\begin{table}[t]
\footnotesize
\centering
\caption{F$_1$ scores of the top two neurons from each network for detecting tokens inside parentheses, and the ranks of the top neuron according to our intrinsic unsupervised methods. 
}
\begin{tabular}{l|lllll||l|lllll}
\toprule
 Neuron  &  1st  & 2nd &  \texttt{Max}  &  \texttt{Min}  & \texttt{Reg}   & Neuron  &  1st  & 2nd &  \texttt{Max}  &  \texttt{Min}  & \texttt{Reg}  \\
 \midrule
en-es-1:232  & 0.59 & 0.3 & 14 & 44 & 26 & en-ar-3:331  & 0.59 & 0.35 & 17 & 92 & 49\\
en-es-2:208  & 0.72 & 0.26 & 8 & 43 & 21 & en-ru-1:259  & 0.64 & 0.33 & 10 & 47 & 44\\
en-es-3:47  & 0.57 & 0.29 & 11 & 34 & 23 & en-ru-2:23  & 0.71 & 0.26 & 10 & 72 & 31\\
en-fr-1:499  & 0.6 & 0.27 & 37 & 41 & 14 & en-ru-3:214  & 0.65 & 0.32 & 25 & 67 & 114\\
en-fr-2:361  & 0.61 & 0.35 & 28 & 44 & 60 & en-zh-1:49  & 0.58 & 0.44 & 5 & 85 & 63\\
en-fr-3:253  & 0.37 & 0.35 & 140 & 122 & 68 & en-zh-2:159  & 0.76 & 0.38 & 5 & 47 & 37\\
en-ar-1:383  & 0.38 & 0.36 & 119 & 195 & 228 & en-zh-3:467  & 0.54 & 0.32 & 5 & 59 & 47\\
en-ar-2:166  & 0.63 & 0.25 & 4 & 117 & 67 &  &  &  &  &  &  \\ 
\bottomrule
\end{tabular}
\label{tab:paren-stats}
\vspace{-7pt}
\end{table}

\paragraph{Parentheses} 
Table~\ref{tab:paren-stats} 
 shows top neurons from each model 
 for predicting that tokens are inside/outside of parentheses, quotes, or brackets, estimated by a GMM model. 
 Often, 
 the parentheses neuron is unique (low scores for the 2nd 
 best neuron), suggesting that this property tends to be relatively localized. 
  Generally, neurons that detect 
parentheses were ranked highly in most  
models by the 
\texttt{MaxCorr} method, 
indicating that they capture important patterns in multiple networks.

The next 
figure 
visualizes
the most predictive neuron in an English-Spanish model. It activates positively (red) 
inside parentheses and negatively (blue) 
outside. 
Similar neurons were found in 
RNN 
language models~\citep{karpathy2015visualizing}. 
Next we consider more complicated linguistic properties.

 \begin{figure}[h]
\centering
\includegraphics[width=0.7\textwidth]{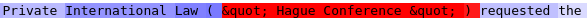}
    \label{fig:paren-vis}
\end{figure}

 \begin{table}[t]
\small
\centering
\caption{Strongest correlations in all models relative to a tense neuron in an English-Arabic model.}
\begin{tabular}{ll|ll|ll}
\toprule
Arabic & 0.66, 0.57  & French & -0.69, -0.58, -0.48 & Chinese & -0.51, -0.30, -0.18\\
Spanish & 0.56, 0.36, 0.22 & Russian & -0.50, -0.39, -0.29 & English & -0.33, -0.19, -0.03\\
\bottomrule
\end{tabular}
\label{tab:max-corr-tense}
\vspace{-5pt}
\end{table}

\paragraph{Tense}
We annotated the test data for verb 
tense (with Spacy) and trained a GMM model to predict tense from neuron activations. 
The following figure 
shows activations of a top-scoring neuron (0.56 F$_1$) from the English-Arabic model on the first 5 test sentences. It tends to activate positively (red color) 
on present tense 
(``recognizes'', ``recalls'', ``commemorate'') and negatively (blue color) on past tense 
(``published'', ``disbursed'', ``held''). 
These 
results are obtained with a charCNN representation, which is sensitive to common suffixes like ``-ed'', ``-es''. However,  this neuron also detects irregular past tense verbs like ``held'', suggesting that it captures 
context in addition to sub-word information.  
The neuron also makes some mistakes by activating weakly positively on nouns ending with ``s'' (``videos'', ``punishments''), presumably because it gets confused with the 3rd person present tense. 

\begin{figure}[h!]
\centering
\includegraphics[width=\linewidth]{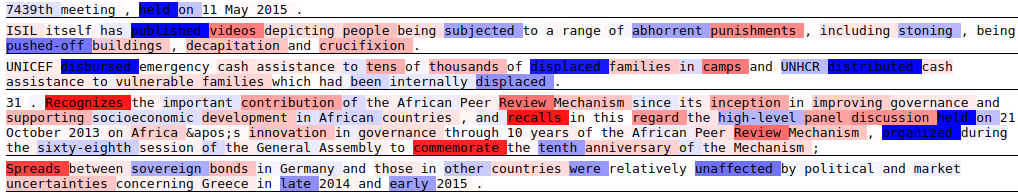}
\label{fig:vis-tense}
\end{figure}

Table~\ref{tab:max-corr-tense}  
shows 
correlations of neurons 
most correlated with this tense neuron, according to  
\texttt{MaxCorr}. 
All these neurons are highly predictive of tense: all are in the top 5 
and 9 out of 15 (non-auto-encoder)  neurons  have the highest F$_1$ score for predicting tense. 
The auto-encoder English 
models are an exception, exhibiting much lower correlations with the English-Arabic tense neuron. This suggests that tense 
emerges in a ``real'' NMT model, but not in an auto-encoder 
that only learns to copy. 
Interestingly, 
English-Chinese models have somewhat lower correlated neurons with the tense neuron, 
possibly due to
the lack of explicit tense marking in Chinese. 
The 
encoder does not need to pay as much attention to tense when generating representations for 
the decoder.

\paragraph{Other Properties}

We found many more linguistic properties by visualizing top neurons ranked by our methods, especially with \texttt{MaxCorr}. We briefly mention some of these here and provide more details and quantitative results in the appendix. 
We found neurons that activate on numbers, dates, adjectives, plural nouns, auxiliary verbs, and more.
We also investigated noun phrase segmentation, a compositional property above the word level, 
and found high-scoring neurons (60-80\% accuracy) in every network. Many of these neurons were ranked highly by the \texttt{MaxCorr}  method. 
In contrast, other 
methods did not rank such neurons very highly. 
See Table~\ref{tab:np-stats} in the appendix for the full results.

Some neurons have quite complicated behavior. 
For example, when visualizing neurons highly ranked by \texttt{MaxCorr} we found a neuron that activates on numbers in the beginning of a sentence, but not elsewhere 
(see Figure~\ref{fig:appendix-number-init} in the appendix).
It would be difficult to conceive of a supervised prediction task which would capture this behavior a-priori, without knowing what to look for.  Our supervised methods are flexible enough to find any neurons deemed important by the NMT model, without  constraining the analysis to properties for which we have supervised annotations.

\vspace{-5pt}
\section{Controlling Translations} \label{sec:control}
\vspace{-7pt}

In this section, 
 we explore a potential benefit of finding important neurons with linguistically meaningful properties: controlling the translation output. This may be important for mitigating biases in neural networks. 
For instance, gender stereotypes are often reflected in automatic translations, as the following motivating examples from Google Translate demonstrate.\footnote{For more biased examples, see \url{mashable.com/2017/11/30/google-translate-sexism}.}  
 \\ 
\parbox{.5\linewidth}{
\eenumsentence{
\item o bir doctor \label{ex:tu-doctor}
\item he is a doctor \label{ex:en-doctor}
}
}
\parbox{.5\linewidth}{
\eenumsentence{
\item o bir hem\c{s}ire \label{ex:tu-nurse}
\item she is a nurse \label{ex:en-nurse}
}
}
The Turkish sentences (\ref{ex:tu-doctor}, \ref{ex:tu-nurse}) have no gender information---they can refer to either 
male or 
female. But the MT system is biased to think that doctors are usually men and nurses are usually women, so its generated translations (\ref{ex:en-doctor}, \ref{ex:en-nurse}) represent these biases.

We conjecture that if a given 
neuron matters to the 
model, then we can control the translation in predictable ways by modifying its  
activations. 
To do this, we first encode the source sentence  as usual. Before decoding, we set the activation of a particular neuron in the encoder state 
to a 
value $\alpha$, which is a function of the mean activations over a particular property (defined below). 
To evaluate our ability to control the translation, we design the following protocol: 
\begin{enumerate}[leftmargin=*,itemsep=0pt,topsep=3pt]
\item[1.] Tag the source and target sentences in the development set with a desired property, such as gender (masculine/feminine). We use Spacy for these tags.
\item[2.] Obtain word alignments for the development set with using 
an  
alignment model trained on 
2 million 
sentences of the UN data. We use 
\texttt{fast\_align}~\citep{N13-1073}  with default settings. 
\item[3.] For every neuron in the encoder, predict the target property on the word aligned to its source word activations using a supervised GMM model.\footnote{This is different from our results in the previous section, where we predicted a source-side property, because here we seek 
neurons that are predictive of target-side properties to facilitate controlling the translation.} 
\item[4.] For every word having 
a desired property, modify the source activations of the top $k$ neurons found in step 3, 
and generate a modified translation. The modification value  is defined as \mbox{$\alpha = \mu_1 + \beta (\mu_1 - \mu_2) $,} where $\mu_1$ and $\mu_2$ are 
mean activations of the property we modify from and to, respectively (e.g. modifying gender from masculine to feminine), and $\beta$ is a hyper-parameter. 
\item[5.] Tag the output translation and word-align it to the source. Declare \textit{success} if the source word was aligned to a target word with the desired property value (e.g. feminine). 
\end{enumerate}

\subsection{Results}

\begin{figure}[t]
\centering
\begin{subfigure}[b]{0.32\textwidth}
\includegraphics[width=\linewidth]{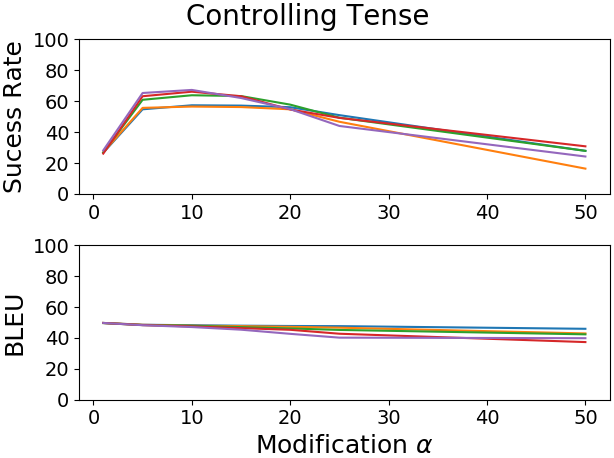}
\caption{Tense}
\label{fig:control-tense}
\end{subfigure} \hfill
\begin{subfigure}[b]{0.32\textwidth}
\includegraphics[width=\linewidth]{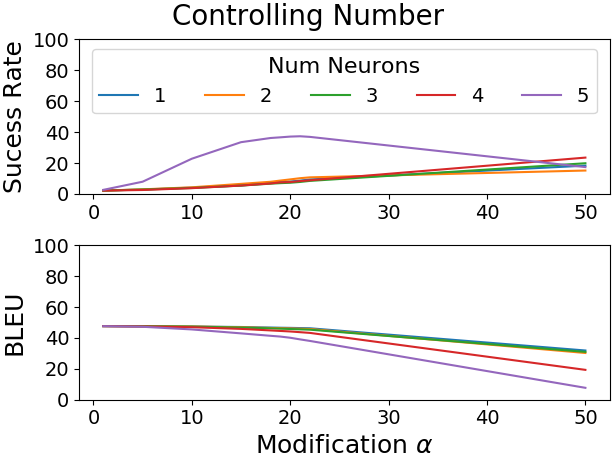}
\caption{Number}
\label{fig:control-number}
\end{subfigure} \hfill
\begin{subfigure}[b]{0.32\textwidth}
\includegraphics[width=\linewidth]{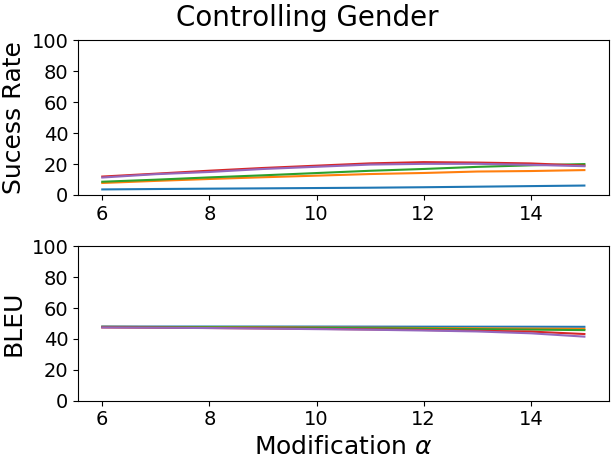}
\caption{Gender}
\label{fig:control-gender}
\end{subfigure} 
\caption{Success rates and BLEU scores for controlling NMT by modifying neuron activations.}
\label{fig:control}
\end{figure}

Figure~\ref{fig:control} shows translation control results 
in an English-Spanish model. We report success rate---the percentage of cases where the word was aligned to a target word with the desired property--and the effect on BLEU scores, when varying 
$\alpha$. 
Our tense control results are the most successful, with up to 67\% success rate for changing past-to-present. Modifications generally degrade BLEU, but the loss at the best success rate is not large (2 BLEU points). We provide more tense results in Appendix~\ref{app:control}. 

Controlling other properties seems more difficult, with the best  success rate for controlling number at 37\%, using the 5 top number neurons. Gender is the most difficult to control, with a 21\% success rate using the 5 top neurons. Modifying even more neurons did not help. 
We conjecture that these properties are more distributed than tense, which makes controlling them more difficult. Future work can explore more sophisticated methods for controlling multiple neurons simultaneously.

\subsection{Example translations}
We provide examples of controlling translation of number, gender, and tense. 
While these are cherry-picked examples, they illustrate that the controlling procedure can work in multiple properties and languages. 
Appendix~\ref{sec:appendix-control} discusses these examples and language-specific behaviors in more detail. 

\paragraph{Number}

Table~\ref{tab:manipulate-number} shows translation 
control results 
for a number neuron from 
an English-Spanish model, which 
activates negatively/positively on plural/singular nouns. 
The translation 
changes from plural to singular
as we increase the modification $\alpha$.
We notice that using too high $\alpha$ values yields 
nonsense translations, but with correct number: 
transitioning from 
  the plural adjective \textit{particulares} (``particular'') 
  to 
  the singular adjective \foreignlanguage{spanish}{\textit{útil}} (``useful''), with valid translations in between. 

\paragraph{Gender}
Table~\ref{tab:manipulate-gender} shows examples of controlling gender translation for a gender neuron from the same 
model, which activates negatively/positively on masculine/feminine nouns. The translations 
change from masculine to feminine synonyms as we increase the modification $\alpha$. Generally, we found it difficult to control gender, as also suggested by the relatively low success rate.

\begin{table}[t]
\footnotesize
\centering
\caption{Examples for controlling translation by modifying activations of different neurons on the \textit{italicized} source words.
$\alpha$ = modification value (--, no modification). }
\begin{subtable}[t]{\linewidth}
\footnotesize
\centering
\caption{Controlling number when translating ``The interested \textit{parties}'' to Spanish. 
}
\begin{tabular}{lll | lll }
\toprule
  \multicolumn{1}{c}{$\alpha$} & \multicolumn{1}{c}{Translation} & \multicolumn{1}{c|}{Num} &  \multicolumn{1}{c}{$\alpha$} & \multicolumn{1}{c}{Translation} & \multicolumn{1}{c}{Num} \\ 
\midrule
 -1 & abiertas particulares  & pl. & 0.125 & La parte interesada & sing. \\ 
 -0.5 & Observaciones interesadas & pl. &  0.25 & Cuestion interesada & sing. \\ 
 -0.25, -0.125, 0 & Las partes interesadas & pl. &  0.5, 1 & \foreignlanguage{spanish}{Gran útil} & sing. \\ 
\bottomrule
\end{tabular}
\label{tab:manipulate-number}
\vspace{3pt}
\end{subtable}
\begin{subtable}[t]{\linewidth}
\footnotesize
\centering
\caption{Controlling gender when 
translating ``The interested \textit{parties}'' (left) and ``\textit{Questions} relating to information'' (right) to Spanish. 
}
\begin{tabular}{p{1.31cm}ll | p{1.35cm}ll}
\toprule
\multicolumn{1}{c}{$\alpha$} & \multicolumn{1}{c}{Translation} & \multicolumn{1}{c|}{Gen} &  \multicolumn{1}{c}{$\alpha$} & \multicolumn{1}{c}{Translation} & \multicolumn{1}{c}{Gen} \\ 
\midrule
-0.5, -0.25 & Los partidos interados & ms.  & -1 & \foreignlanguage{spanish}{Temas relativos a la información} & ms. \\
0, 
0.25 
& Las partes interesadas & fm. & -0.5, 
0, 
0.5 
& \foreignlanguage{spanish}{Cuestiones relativas a la información} & fm.  \\ 
\bottomrule
\end{tabular}
\label{tab:manipulate-gender}
\vspace{3pt}
\end{subtable}
\begin{subtable}[t]{\linewidth}
\footnotesize
\centering
\caption{Controlling  tense when translating ``The committee \textit{supported} the efforts of the authorities''. 
}
\begin{tabular}{l|l|p{8.2cm}|p{2cm}}
\toprule
 &  
 \multicolumn{1}{c|}{$\alpha$} & \multicolumn{1}{c|}{Translation}  & \multicolumn{1}{c}{Tense} \\
\midrule
Arabic &  
--/+10 &   \multicolumn{1}{R}{\foreignlanguage{arabic}{وأيدت/وتؤيد اللجنة \}جهود/الجهود التي تبذلها\{ السلطات}}  & \multicolumn{1}{|l}{past/present} \\
\midrule
\multirow{1}{*}{French} & 
--/-20 &  \foreignlanguage{french}{Le Comité \underline{a appuyé}/\underline{appuie} les efforts des autorités}  & past/present \\ 
\midrule
\multirow{1}{*}{Spanish} & 
--/-3/0 & \foreignlanguage{spanish}{El Comité \underline{apoyó}/\underline{apoyaba}/\underline{apoya} los esfuerzos de las autoridades} & past/impf./present \\
\midrule
\multirow{1}{*}{Russian} & 
--/-1 & \foreignlanguage{russian}{Комитет \underline{поддержал}/\underline{поддерживает} усилия властей} &  past/present \\
\midrule
Chinese & 
--/-50 & \begin{CJK}{UTF8}{gbsn} 委员会 支持 当局 的 努力 \end{CJK} / \begin{CJK}{UTF8}{gbsn} 委员会 \underline{正在} 支持 当局 的 努力 \end{CJK}  & untensed/present \\
\bottomrule
\end{tabular}
\label{tab:manipulate-past-present}
\end{subtable}
\end{table}


\paragraph{Tense}

Table~\ref{tab:manipulate-past-present} shows examples of controlling tense 
when translating from English to five target languages. 
In all language pairs, we are able to change the translation
from past to present by modifying the activation of the tense neurons from the previous section (Table~\ref{tab:max-corr-tense}).  
  In Spanish, 
  we find 
  a transition from past  
  to imperfect 
  to present. 
Interestingly,
  in Chinese, we had to use a fairly large $\alpha$ value (in absolute terms), 
  consistent with the fact that tense is not usually marked in Chinese.

\section{Conclusion}
We developed unsupervised
methods for finding important neurons in NMT, and evaluated how these neurons impact translation quality.
We analyzed several linguistic properties that are captured by individual neurons using quantitative prediction tasks and qualitative visualizations.  
We also designed a protocol for controlling translations 
by modifying neurons 
that capture 
desired  
properties.

Our analysis can be extended to other NMT components (e.g. the decoder) and architectures~\cite{pmlr-v70-gehring17a,NIPS2017_7181},  
as well as 
other 
tasks. 
We believe that more work should be done to analyze the spectrum of localized vs.\ distributed information in neural language representations.  
We would also like to develop more sophisticated ways to control translation output, for example by modifying representations in variational NMT architectures~\citep{D16-1050,su2018variational}.

\section*{Acknowledgments}
This research was carried out in collaboration between the HBKU Qatar Computing Research Institute (QCRI) and the MIT Computer Science and Artificial Intelligence Laboratory (CSAIL).

\bibliography{naaclhlt2018,iclr2019}
\bibliographystyle{iclr2019_conference}


\appendix


\section{Additional Results and Visualizations} \label{app:results}
\subsection{Noun Phrase Segmentation}
Table~{\ref{tab:np-stats} shows the top neurons from each network by accuracy when classifying interior, exterior, or beginning of a noun phrase. The annotations were obtained with Spacy. 
We found high-scoring neurons (60-80\% accuracy) in every network. Many of these neurons were ranked highly by the \texttt{MaxCorr} ranking methods. 
In contrast, other correlation methods did not rank such neurons very highly. 
Thus there is correspondence between a high rank by our intrinsic unsupervised measure \texttt{MaxCorr} and the neuron's  capacity to predict external annotation.

\begin{table}[h]
\centering
\caption{Top neuron from each network by accuracy for classifying interior, exterior, or beginning of a noun phrase, as well as ranking of these neurons by our intrinsic unsupervised measures.}
\begin{tabular}{l | l | l | l | l}
\toprule
& & \multicolumn{3}{c}{Rank} \\ 
Neuron & Accuracy & \texttt{MaxCorr} & \texttt{MinCorr} & \texttt{LinReg} \\
\midrule
en-es-1:221 & 0.79 & 8 & 54 & 57\\
en-es-2:158 & 0.77 & 11 & 59 & 65\\
en-es-3:281 & 0.73 & 24 & 39 & 111\\
en-fr-1:111 & 0.77 & 12 & 62 & 122\\
en-fr-2:85 & 0.73 & 32 & 45 & 86\\
en-fr-3:481 & 0.76 & 13 & 65 & 133\\
en-ar-1:492 & 0.69 & 32 & 59 & 161\\
en-ar-2:190 & 0.75 & 48 & 80 & 90\\
en-ar-3:288 & 0.69 & 25 & 84 & 157\\
en-ru-1:38 & 0.66 & 35 & 54 & 158\\
en-ru-2:130 & 0.67 & 34 & 64 & 134\\
en-ru-3:78 & 0.67 & 159 & 106 & 123\\
en-zh-1:427 & 0.64 & 22 & 75 & 240\\
en-zh-2:199 & 0.65 & 187 & 216 & 232\\
en-zh-3:28 & 0.68 & 63 & 32 & 42\\
\bottomrule
\end{tabular}
\label{tab:np-stats}
\vspace{-10pt}
\end{table}

\subsection{Controlling translations} \label{app:control}
We provide additional translation control results. 
Table~\ref{tab:results-control-tense} shows the tense results using the best modification value from Figure~\ref{fig:control-tense}.
We report the number of times the source word was aligned to a target word which is past or present, or 
to multiple words that include both or neither of these tenses. The success rate is the percentage of cases where the word was aligned to a target word with the desired tense.  
By modifying the activation of only one neuron (the most predictive one), we were able to change the translation from past to present in 67\% of the times and vice-versa in 49\% of the times. In many other cases, the tense was erased, that is, 
the modified source word was not aligned to any tensed word, 
which is a partial success. 

\begin{table}[h]
\footnotesize
\centering
\caption{Results for controlling tense.}
\begin{tabular}{l|rrrr|c}
\toprule
\backslashbox{From}{To} 
& Past & Present & Both & Neither & Success Rate \\
\midrule
Past     &  85    &  820     &  9    &  311     &  67\% \\ 
Present   &  1586   &  256     &  30    &  1363    &  49\% \\ 
\bottomrule
\end{tabular}
\label{tab:results-control-tense}
\end{table}

\clearpage

\subsection{Visualizations}
Here we provide additional visualizations of neurons capturing different linguistic properties. 

\paragraph{Noun phrases}
We visualize the top scoring neuron (79\%) from an English-Spanish model in 
Figure~\ref{segmentation-vis}.  Notice how the  neuron activates positively (red color) on the first word in the noun phrases, but negatively (blue color) on the rest of the noun phrase (e.g. ``Regional'' in ``Regional Service Centre''). 

\begin{figure}[h]
\centering
	\includegraphics[width=0.7\textwidth]{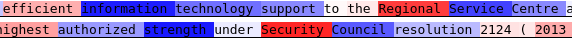}
    \caption{Visualization of a neuron from an English-Spanish model that 
    activates positively (red color) on the first word in the noun phrase and negatively (blue) on the following words. 
    }
    \label{segmentation-vis}
\end{figure}

\paragraph{Dates and Numbers}

Figure~\ref{fig:appendix-numbers} shows activations of neurons capturing dates and numbers. These neurons were ranked highly (top 30) by \texttt{MaxCorr} when ranking an English-Arabic model trained with charCNN representations. We note that access to character information leads to many neurons capturing sub-word information such as years (4-digit numbers).  
The first neuron is especially sensitive to month names (``May'', ``April''). The second neuron is an approximate year-detector: it is sensitive to years (``2015'') as well as other tokens with four digits (``7439th'', ``10.15''). 

\begin{figure}[h]
\centering
\begin{subfigure}{0.6\textwidth}
\includegraphics[width=\linewidth]{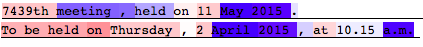}
\caption{Month neuron}
\end{subfigure}
\begin{subfigure}{0.6\textwidth}
\includegraphics[width=\linewidth]{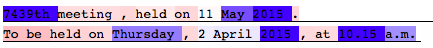}
\caption{Approximate ``year'' neuron}
\end{subfigure}
\caption{Neurons capturing dates and numbers.}
\label{fig:appendix-numbers}
\end{figure}


\paragraph{List items}
Figure~\ref{fig:appendix-number-init} shows an interesting case of a neuron that is sensitive to the appearance of two properties simultaneously:  position in the beginning of the sentence and  number format. 
Notice that it activates strongly (negatively) on numbers when they open a sentence but not in the middle of the sentence. Conversely, it does not activate strongly on non-number words that open a sentence. This neuron aims to capture patterns of opening list items.

\begin{figure}[h]
\centering
\includegraphics[width=\linewidth]{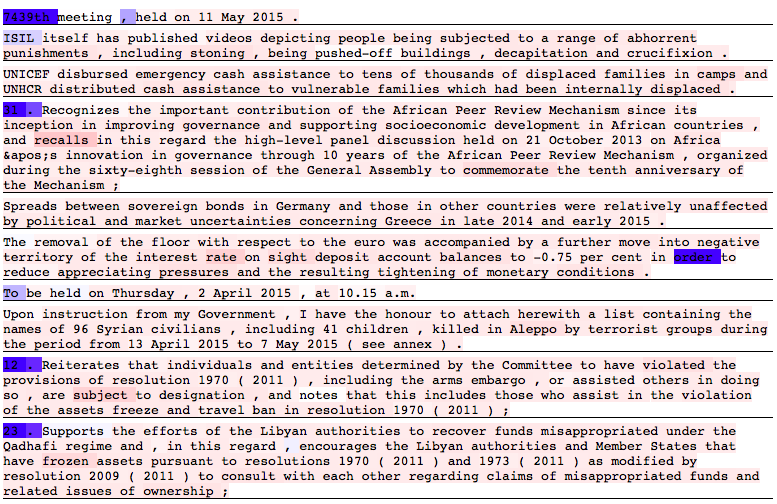}
\caption{A neuron that activates on numbers in the beginning of sentences. The first 10 sentences in the test set are shown.}
\label{fig:appendix-number-init}
\end{figure}

\clearpage

\section{Examples for controlling translations} \label{sec:appendix-control}

We provide here a more detailed discussion of the example translations from Section~\ref{sec:control}.

\paragraph{Number}

Table~\ref{tab:manipulate-number} shows translation 
control results 
for a number neuron from 
an English-Spanish model, which 
activates negatively/positively on plural/singular nouns. 
The table shows how the translation of the phrase ``The interested \textit{parties}'' changes from plural to singular, 
as we increase the modification $\alpha$.
Notice that using too high $\alpha$ values yields 
nonsense translations, but with correct number: high negative values give the plural adjective \textit{particulares} (``particular''); 
high positive values give the singular adjective \foreignlanguage{spanish}{\textit{útil}} (``useful''). In between, we see a nice transition between plural and singular translations. 
Interestingly, the translations exhibit correct agreement between the modified noun and its adjectives and determines, e.g., \textit{Las partes interesadas} vs.\ \textit{La parte interesada}. This is probably due to the strong language model in the decoder.

\paragraph{Gender}
Table~\ref{tab:manipulate-gender} 
shows examples of controlling gender translation for a gender neuron from the same English-Spanish 
model, which activates negatively/positively on masculine/feminine nouns,  for two phrases: ``The interested \textit{parties}'' and ``\textit{Questions relating to information}''. 
The translations of ``parties'' and ``questions'' 
change from masculine to feminine synonyms as we increase the modification $\alpha$.

\paragraph{Tense}

Table~\ref{tab:manipulate-past-present} shows  tense 
control results for the sentence ``The committee \textit{supported} the efforts of the authorities''. 
In all cases, we are able to change the translation of ``supported'' 
from past to present by modifying the activation of the tense neurons  (Table~\ref{tab:max-corr-tense}). 
Occasionally, modifying the neuron 
activation on a single word leads to a change in phrasing; in Arabic the translation changes to ``the efforts that the authorities invest''. 
In Spanish, 
we find a transition from past (\textit{\foreignlanguage{spanish}{apoyó}}) to imperfect (\textit{apoyaba})  to present (\textit{apoya}). 
Finally,  
note that in Chinese, we had to use a fairly large $\alpha$ value (in absolute terms) to generate a manipulation. This is consistent with the fact that tense is not usually marked in Chinese. In fact, our modification generates  a Chinese expression (\begin{CJK}{UTF8}{gbsn}正在\end{CJK}) that is used to express an action in progress, similar to English ``-ing'', resulting in the meaning ``is supporting''. 

\clearpage

\section{A Catalog of Top Ranked Neurons}
In order to illustrate the range of linguistic phenomena captures by individual neurons, we provide here a list of the top 20 neurons (or projected directions, in the case of SVCCA) found by each of our methods, for an example English-Spanish model. 
For each neuron, we give the 
percentage of variance that is eliminated by conditioning on position in the sentence or identity of the current token. 
We also comment on  what properties each neuron appears to capture, based on visualizations of neuron activation.
Where possible, we give F$_1$ scores of a GMM model  predicting certain properties such as detecting noun phrase segmentation, parenthetical phrases, adjectives, and plural nouns. Annotations are obtained with Spacy: \url{https://spacy.io}.



\begin{table}[h]
\centering
\caption{Top 20 ranked neurons by \texttt{MaxCorr}.}
\begin{tabularx}{\textwidth}{l | l | l | X}
\toprule
Neuron & Position & Token & Comments\\
\midrule
464 & 92\% & 10\% & Position.\\
342 & 88\% & 7.9\% & Position.\\
260 & 0.71\% & 94\% & Conjunctions: "and", "or", "well", "addition".\\
49 & 11\% & 6.1\% & Activates for several words after "and" or "or".\\
124 & 77\% & 48\% & Position.\\
394 & 0.38\% & 22\% & Noun phrase segmentation. 3rd-best F1-score (0.56) for finding interiors of noun phrases. 8th-best IOB accuracy (0.64).\\
228 & 0.38\% & 96\% & Unknown: "'s", "the", "this", "on", "that".\\
133 & 0.14\% & 87\% & Adjective detector. Best F1-score (0.56) for finding adjectives.\\
221 & 1\% & 30\% & Noun phrase segmentation. Best F1-score for finding interiors (0.72) and for finding beginnings (0.66). Best IOB accuracy (0.80).\\
90 & 0.49\% & 28\% & Noun phrase segmentation. Best F1-score (0.59) for finding beginnings on noun phrases. Second-best IOB accuracy (0.73).\\
383 & 67\% & 6.5\% & Position.\\
494 & 3.5\% & 96\% & Punctuation/conjunctions: ",", ";", "Also", "also", "well".\\
120 & 0.094\% & 84\% & Plural noun detector. Best F1-score (0.87) for retrieving plural nouns.\\
269 & 0.1\% & 80\% & Spanish noun gender detector. Very positive for "islands", "activities", "measures" -- feminine. Very negative for "states", "principles", "aspects" -- masculine.\\
232 & 0.63\% & 31\% & Parentheses. Best F1-score (0.60) for retrieving tokens inside parentheses/quotes/brackets.\\
332 & 0.13\% & 83\% & Unknown.\\
324 & 0.18\% & 81\% & Unknown.\\
210 & 0.61\% & 45\% & Date detector. Third-best F1-score (0.39) for retrieving tokens inside dates.\\
339 & 0.48\% & 39\% & Activates for a verb and also surrounding inflection words/auxiliary verbs.\\
139 & 0.86\% & 93\% & Punctuation/conjunctions: ",", ".", "--". \\ 
\bottomrule
\end{tabularx}
\end{table}


\begin{table}
\centering
\caption{Top 20 ranked neurons by \texttt{MinCorr}.}
\begin{tabularx}{\textwidth}{l | l | l | X}
\toprule
Neuron & Position & Token & Comments\\
\midrule
342 & 88\% & 7.9\% & Position.\\
464 & 92\% & 10\% & Position.\\
260 & 0.71\% & 94\% & Conjunctions: "and", "or", "well", "addition".\\
383 & 67\% & 6.5\% & Position.\\
250 & 63\% & 6.8\% & Position.\\
124 & 77\% & 48\% & Position.\\
485 & 64\% & 10\% & Position.\\
480 & 70\% & 12\% & Position.\\
154 & 63\% & 15\% & Position.\\
139 & 0.86\% & 93\% & Punctuation/conjunctions: ",", ".", "--", "alia".\\
20 & 60\% & 9.2\% & Position.\\
494 & 3.5\% & 96\% & Punctuation/conjunctions: ",", ";", "also", "well".\\
199 & 67\% & 6\% & Position.\\
126 & 42\% & 9.4\% & Unknown.\\
348 & 50\% & 13\% & Position.\\
46 & 48\% & 8.6\% & Unknown.\\
196 & 60\% & 8.5\% & Position.\\
367 & 0.44\% & 89\% & Prepositions: "of", "or", "United", "de".\\
186 & 1.6\% & 69\% & Conjunctions: "also", "therefore", "thus", "alia".\\
244 & 54\% & 15\% & Position.\\
\bottomrule
\end{tabularx}
\end{table}


\begin{table}
\centering
\caption{Top 20 ranked neurons by \texttt{LinReg}.}
\begin{tabularx}{\textwidth}{l | l | l | X}
\toprule
Neuron & Position & Token & Comments\\
\midrule
464 & 92\% & 10\% & Position.\\
260 & 0.71\% & 94\% & Conjunctions: "and", "or", "well", "addition".\\
139 & 0.86\% & 93\% & Punctuation/conjunctions: ",", ".", "--", "alia".\\
494 & 3.5\% & 96\% & Punctuation/conjunctions: ",", ";", "also", "well".\\
342 & 88\% & 7.9\% & Position.\\
228 & 0.38\% & 96\% & Possibly determiners: ""s", "the", "this", "on", "that".\\
317 & 1.5\% & 83\% & Indefinite determiners: "(", "one", "a", "an".\\
367 & 0.44\% & 89\% & Prepositions. "of", "for", "United", "de", "from", "by", "about".\\
106 & 0.25\% & 92\% & Possibly determiners: "that", "this", "which", "the".\\
383 & 67\% & 6.5\% & Position.\\
485 & 64\% & 10\% & Position.\\
186 & 1.6\% & 69\% & Conjunctions. "also", "therefore", "thus", "alia".\\
272 & 2\% & 73\% & Tokens that mean "in other words": "(", "namely", "i.e.", "see", "or".\\
124 & 77\% & 48\% & Position.\\
480 & 70\% & 12\% & Position.\\
187 & 1.1\% & 87\% & Unknown: "them", "well", "be", "would", "remain".\\
201 & 0.14\% & 73\% & Tokens that mean "regarding": "on", "in", "throughout", "concerning", "regarding".\\
67 & 0.27\% & 71\% & Unknown: "united", "'s", "by", "made", "from".\\
154 & 63\% & 17\% & Position.\\
72 & 0.32\% & 89\% & Verbs suggesting equivalence: "is", "was", "are", "become", "constitute", "represent". \\
\bottomrule
\end{tabularx}
\end{table}


\begin{table}
\caption{Top 20 ranked directions by \texttt{SVCCA}.}
\begin{tabularx}{\textwidth}{l | l | X}
\toprule
Position & Token & Comments\\
\midrule
86\% & 26\% & Position\\
1.6\% & 90\% & Detects "the".\\
7.5\% & 85\% & Conjunctions: "and", "well", "or".\\
20\% & 79\% & Determiners: "the", "this", "these", "those".\\
1.1\% & 89\% & Possibly conjunctions: negative for "and", "or", "nor", positive for "been", "into", "will".\\
10\% & 76\% & Punctuation/conjunctions: positive for ",", ";", "." "--", negative for "and".\\
30\% & 57\% & Possibly verbs: "been", "will", "be", "shall".\\
24\% & 55\% & Possibly date detector.\\
23\% & 60\% & Possibly adjective detector.\\
18\% & 63\% & Unknown.\\
4.5\% & 88\% & Punctuation: ".", ",", ";"\\
9.8\% & 69\% & Forms of "to be": "is", "will", "shall", "would", "are".\\
1.7\% & 77\% & Combined dates/prepositions/parentheses: negative for "in", "at", ".", positive for dates and in quotes/parentheses/brackets. Noisy.\\
16\% & 25\% & Activates for a few words after "and".\\
14\% & 63\% & Possibly plural noun detector.\\
0.8\% & 73\% & Spanish noun gender detector.\\
11\% & 61\% & Possibly singular noun detector.\\
13\% & 58\% & Possibly possessives: "its", "his", "their".\\
1.4\% & 73\% & Spanish noun gender detector.\\
5.6\% & 53\% & Unknown.\\
\bottomrule
\end{tabularx}
\end{table}

\end{document}